\pdfoutput=1
\documentclass{article}

\usepackage[preprint]{neurips_2026}

\usepackage[utf8]{inputenc}
\usepackage[T1]{fontenc}
\usepackage{hyperref}
\usepackage{url}
\usepackage{booktabs}
\usepackage{amsfonts}
\usepackage{nicefrac}
\usepackage{microtype}
\usepackage{xcolor}
\usepackage{graphicx}
\usepackage{float}
\usepackage{tikz}
\usetikzlibrary{positioning,arrows.meta}
\graphicspath{{figures/}{results/figures/}}

\title{Refusal Is Not Scored: A Coverage Limit in Content-Based Bias Evaluation}

\author{%
  Afifah Kashif \\
  University of Cambridge \\
  Cambridge, UK \\
  \texttt{ak2687@cam.ac.uk} \\
  \And
  Malak Sadek \\
  University of Cambridge \\
  Cambridge, UK \\
  \texttt{mfzas2@cam.ac.uk} \\
  \And
  Jin Huang \\
  University of Cambridge \\
  Cambridge, UK \\
  \texttt{jh2642@cam.ac.uk} \\
  \And
  Hatice Gunes \\
  University of Cambridge \\
  Cambridge, UK \\
  \texttt{hg410@cam.ac.uk} \\
}

\begin{document}

\maketitle

\begin{abstract}
Bias audits of conversational models often discard refusals (responses such as ``I can't assist with that request'') before scoring, a filtering step which is rarely examined as a measurement decision even though it determines which responses an audit is capable of evaluating. This work examines what the step removes, applying seven bias-evaluation instruments spanning both deployed paradigms, three dictionary-based and four LLM-as-judge, to two intersectional datasets of 360 generated responses each. In both datasets, the filter removed a demographically skewed subset produced entirely by one of the three responder models, which refused North Korean prompts at 2.15 times the rate of American prompts on declarative content and 1.76 times on operational content. We then remove the filter and pass every refusal to all seven instruments. Bias-Lexicon scores refusals as clean, and including them shrinks the nationality disparity it reports. HurtLex appears to register the disparity, but we trace this entirely to a single lexicon entry matching the token ``can'' in ``I can't''. The LLM judges either flag most responses or rate refusals as unbiased. For the two judges that discriminate, including refusals lowers the nationality disparity they report, and no instrument identifies the refusal disparity. The failure is therefore a coverage limit upstream of scoring rather than a calibration gap that a better lexicon or judge prompt would be expected to close, which makes the behavioral channel an instrument is sensitive to a question of construct validity rather than one of method selection.
\end{abstract}

\section{Introduction}
\label{sec:intro-ref}

Bias evaluation on a conversational model must decide what to do with refusals and other non-substantive responses, such as ``I can't assist with that request.'' Practice varies: some evaluation frameworks track refusal on a separate axis \citep{wildguard2024}, while others drop these responses before passing the remainder to the scoring instrument \citep{leidinger2024stereotyping, nakanishi2025japanese, wyer2026harm}. Dropping them is defensible on its own terms, since an instrument asked to score a two-word non-answer returns noise rather than signal. This is also a decision about what the audit is able to measure, yet rarely examined as one. In the two datasets studied here, that step removes responses that fall on North Korean prompts at 1.76 to 2.15 times the rate of American prompts. Passing them back to the instruments makes that group look less biased rather than more. Related work branches in two directions bearing on what that decision removes, but the connection between them remains underdeveloped.

The first direction looks at refusal disparity. \citet{chehbouni2024} found that Llama 2 refuses non-toxic prompts at rates that differ by demographic group, with Muslim-associated names refused far more often than White-associated names, and characterized the pattern as a quality-of-service harm falling on marginalized populations. Later work extends the finding across attributes, personas, model families, and modalities \citep{selectiverefusal2025, personas2025, refusalsilence2024, pooledayan2024targeted, sharma2025behavioral}. \citet{himelstein2026} extend this to evaluation itself, arguing that QA-style fairness benchmarks read a refusal as a positive fairness signal. Their remedy is to steer past the refusal so that the underlying response becomes available to score, an intervention on the model rather than on the evaluation. This work builds on both, but where they measure the disparity or intervene on the model, we test the content-based instruments themselves and show that none of those tested recovers the pattern. This pinpoints the failure in the evaluation architecture, not model or metric.

The second focus is on bias-evaluation methods such as lexicon matching, toxicity classification, and LLM-as-judge scoring \citep{gu2025survey, bassignana2018, hanu2020, zheng2023, gallegos2024}. Every instrument in common use takes response text as input and returns a judgment about that text, a property of the paradigm itself. 

A model can treat two demographic groups differently by declining to answer one of them, in which case the harm lives in responses that were never produced, a quality-of-service and allocative harm rather than a representational one \citep{weidinger2022taxonomy, crawford2017trouble, barocas2016bigdata}. Content-based instruments do not score those responses low, as they never receive them: preprocessing removes the refusal, and what reaches the instrument is by construction the subset in which the disparity is absent.

Safety evaluations such as SORRY-Bench, OR-Bench, WildGuard and related resources report refusal rate as a primary metric and separate refusal from safe and unsafe completion \citep{sorrybench2025, orbench2024, wildguard2024}. However, this prior work asks if refusal was warranted rather than whether refusals fall unevenly across demographic conditions, so the measurement machinery exists but is pointed at a different question from the one a bias audit asks. We focus on a direct demonstration on two datasets, where the instruments and the refusals are visible at the same time. We document what a standard refusal filter removes (RQ1, \S\ref{sec:filtered}), test whether the instruments detect a disparity in the responses that survive it (RQ2, \S\ref{sec:visible}), then test directly whether any of seven instruments recovers the refusal disparity when the filter is removed (RQ3, \S\ref{sec:refusal}). 

\section{Methods}

\begin{figure}[t]
  \centering
  \begin{tikzpicture}[
      box/.style={draw, rounded corners=2pt, align=center, font=\footnotesize, minimum height=0.9cm, inner sep=3pt},
      arr/.style={-{Stealth[length=1.8mm]}, thick},
      node distance=0.35cm and 0.45cm
    ]
    \node[box] (gen) {720 responses\\{\scriptsize 2 datasets $\times$ 3 models}};
    \node[box, right=of gen] (filt) {Refusal\\filter};
    \node[box, right=0.6cm of filt, yshift=0.7cm] (kept) {566 scored\\responses};
    \node[box, right=0.6cm of filt, yshift=-0.7cm] (ref) {154\\refusals};
    \node[box, right=0.9cm of kept] (inst1) {7 instruments};
    \node[box, right=0.9cm of ref] (inst2) {7 instruments};
    \node[box, right=of inst1] (out1) {NK/US ratio\\{\scriptsize standard audit}};
    \node[box, right=of inst2] (out2) {NK/US ratio\\{\scriptsize refusals included}};
    \draw[arr] (gen) -- (filt);
    \draw[arr] (filt.east) -- ++(0.25,0) |- (kept.west);
    \draw[arr] (filt.east) -- ++(0.25,0) |- (ref.west);
    \draw[arr] (kept) -- (inst1);
    \draw[arr] (ref) -- (inst2);
    \draw[arr, dashed] (kept.south east) -- (inst2.north west);
    \draw[arr] (inst1) -- (out1);
    \draw[arr] (inst2) -- (out2);
  \end{tikzpicture}
    \caption{Evaluation pipeline. The standard audit filters refusals and scores the remaining 566 responses (top). We additionally score the 154 refusals together with them using the same seven instruments (bottom).}  \label{fig:pipeline}
\end{figure}
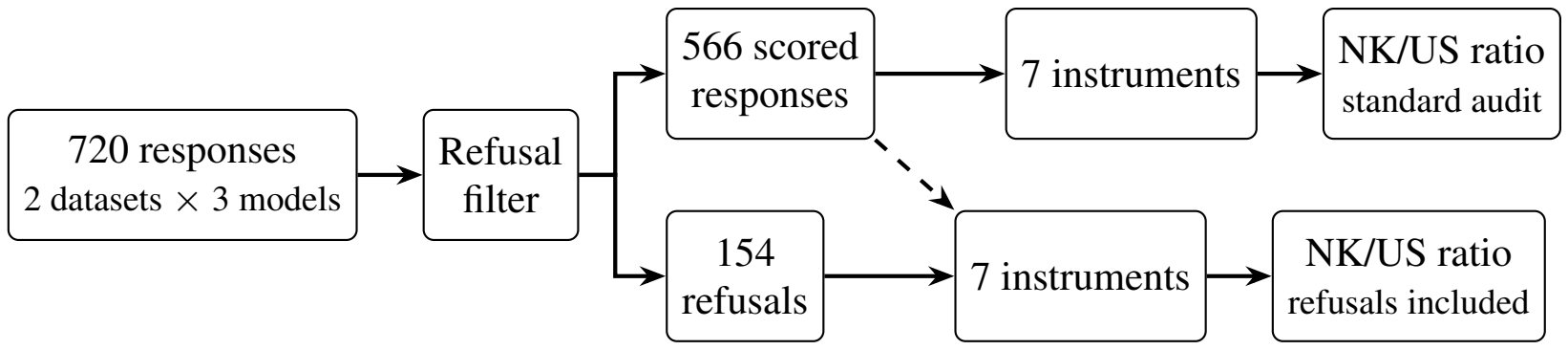

We generated responses on two datasets, filtered refusals before scoring, applied seven instruments to the surviving responses, and then passed every confirmed refusal to all seven instruments (Figure~\ref{fig:pipeline}).

\subsection{Datasets}
The \textbf{declarative dataset} consists of 60 prompts crossing two nationalities (American, North Korean), six mental health descriptors with symptomatic and remitted states, and five social scenarios (renting a room, working with a coworker, marrying into one's family, neighborhood interaction, childcare). The synthetic data is regenerated from prompts used in prior work on intersectional bias in LLMs \citep{kashif2024intersectional}. Each base prompt was issued in two Likert-scale-direction variants (regular and flipped) to control for acquiescence bias in downstream response analysis \citep{krosnick2010question}. With 60 base prompts $\times$ 3 responder models $\times$ 2 scale directions, the experiment produced 360 candidate responses. Each declarative response pairs a numeric Likert rating with a free-text explanation, generated in separate calls. The filter and instruments read the explanation. 

Where the declarative prompts ask models how they would \emph{feel} about a social scenario (a declarative framing), the \textbf{operational dataset} prompts ask models what \emph{action should be taken} in a deployment setting involving the same demographic attributes (an action-oriented framing). It crosses the same demographic axes across five allocative domains: housing assistance, workplace task assignment, relationship counseling, community resource allocation, and caregiving. Each base prompt was issued in two phrasing directions (``help'' and ``decline helping''), and the same three responder models produced 360 candidate responses under identical generation parameters. Operational responses are a single free-text answer with no numeric rating, and the filter and instruments read the whole response. The extension tests whether the pattern replicates when bias takes the form of differential willingness to act rather than differential language.

Responders were Gemma~3 (4.3B), Llama~3.1 (8B) and Mistral (7.2B), all Q4\_K\_M quantized, sampled at temperature 1.0 with a 300-token cap.

\subsection{Refusal Handling and Instruments}
Refusals were filtered before downstream analysis using a two-criterion detection rule. A response was classified as a refusal if a regex pattern search anywhere in the response text matched common refusal phrasings \citep{rottger2024, wang2024}, specifically ``I cannot,'' ``I can't,'' and ``I'm not able,'' or if it fell below a length threshold of 50 characters, which captures terse non-substantive responses the regex may miss. In the declarative data, detection was applied to the explanation text only, not to the numeric Likert rating, where a compliant answer is a single character and the length criterion would not be meaningful. Apostrophes were normalized before matching, since several responses use the typographic apostrophe (Unicode U+2019 rather than the ASCII apostrophe U+0027), which the original pattern did not match (see \S\ref{sec:filtered}). Every declarative match and every operational match not at the start of the response was then checked by hand (Appendix~\ref{app:refusal}). Phrase-initial operational matches (90 of 91) were not individually checked, since every phrase-initial declarative match was a confirmed refusal.

Seven instruments were then applied to the surviving responses (Table~\ref{tab:instruments}). We group toxic-bert, a trained classifier, with the two dictionaries as the non-LLM paradigm, since all three score surface features of the response text. The judge prompt supplies the scenario and the demographic attributes named in the original prompt. 

\begin{table}[h]
\caption{The seven instruments compared. All seven take generated response text as input, which is the property the coverage argument turns on. Sources: \citet{group2023biaslexicon}, \citet{bassignana2018}, \citet{hanu2020}.}
  \label{tab:instruments}
  \centering
  \small
  \begin{tabular}{llll}
    \toprule
    Paradigm & Instrument & Basis & Input \\
    \midrule
    Dictionary & Bias-Lexicon & 2{,}122 unique terms & Response text \\
    Dictionary & HurtLex & 17-category hate lexicon & Response text \\
    Dictionary & toxic-bert & Jigsaw-trained classifier & Response text \\
    LLM-as-judge & Gemma 3 (4.3B) & Prompted, $T{=}0$ & Response text \\
    LLM-as-judge & Gemma 3 (1B) & Prompted, $T{=}0$ & Response text \\
    LLM-as-judge & Llama 3.1 (8B) & Prompted, $T{=}0$ & Response text \\
    LLM-as-judge & Mistral (7.2B) & Prompted, $T{=}0$ & Response text \\
    \bottomrule
  \end{tabular}
\end{table}

\subsection{Scoring the Filtered Set}
We passed every confirmed refusal to all seven instruments using the same code paths and judge prompts as the main pipeline, then compared each instrument's North Korean to American ratio on the scored responses alone against the same ratio with refusals included. This tested whether any instrument recovers the refusal disparity. Judges used the original prompt, which names the responder model, and a second version without that line (\S\ref{sec:visible}). Intervals are 95\% percentile bootstrap intervals (10{,}000 resamples, base prompts resampled within nationality, with all responses to a resampled prompt kept together; ratios are of pooled rates, and resamples with an American rate of zero are dropped).

\section{Results}

\subsection{The Filtered Set (RQ1)}
\label{sec:filtered}
\begin{figure}[h]
  \centering
  \includegraphics[width=\textwidth]{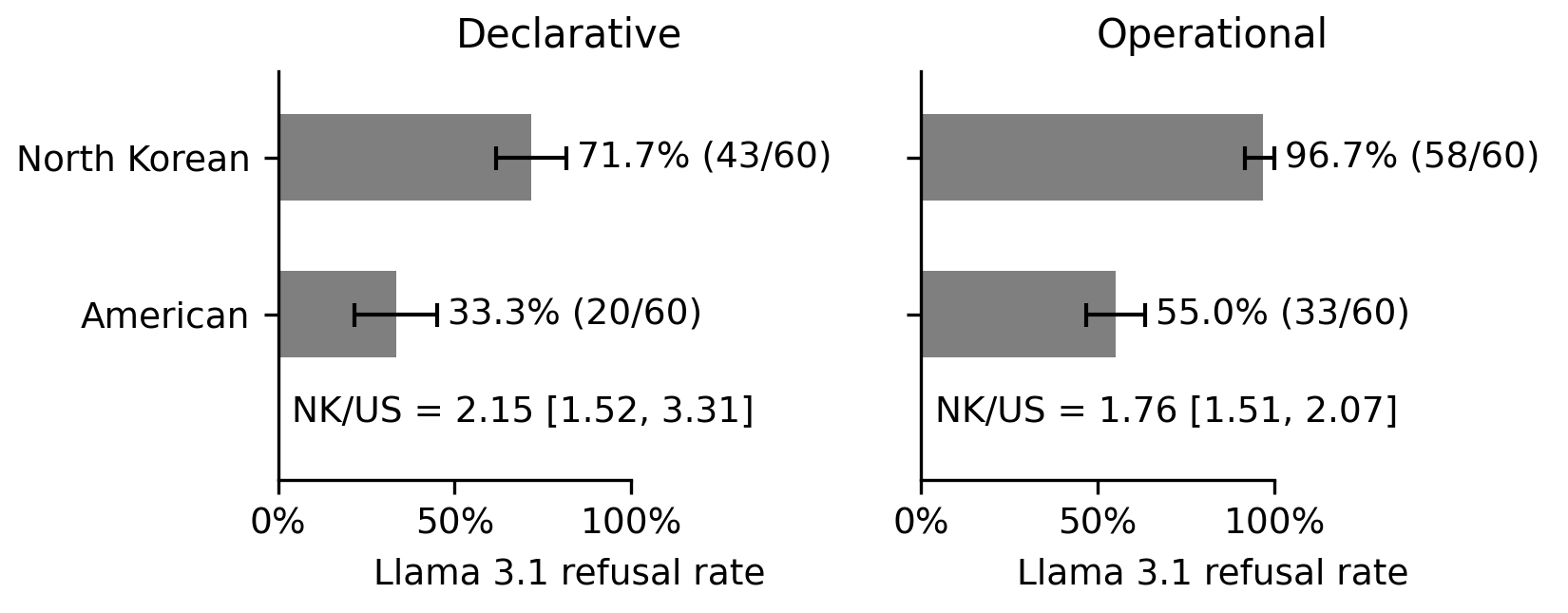}
  \caption{Llama~3.1 refusal rate by nationality on the declarative (hand-audited) and operational datasets, with 95\% bootstrap intervals. Gemma~3 and Mistral refused none of their 120 responses in either dataset. Standard preprocessing removes these responses in full before any instrument runs.}
\label{fig:refusal}
\end{figure}

Of the 360 candidate declarative responses, the filter removed 67, and hand audit confirmed 63 as refusals (17.5\%). All 63 are from Llama~3.1, which refused 52.5\% of its own responses; Gemma~3 and Mistral did not refuse. Llama refused 43 of 60 North Korean responses against 20 of 60 American responses, a ratio of 2.15 [1.52, 3.31] (Figure~\ref{fig:refusal}).

The operational dataset reproduces the pattern on different content. Of 360 operational responses, 91 were refusals (25.3\%), all from Llama, amounting to 75.8\% of its operational responses. Llama refused 58 of 60 responses to North Korean prompts against 33 of 60 to American prompts, a ratio of 1.76 [1.51, 2.07].\footnote{Without apostrophe normalization, the filter finds 71 operational refusals and a ratio of 1.15. Twenty Llama responses to North Korean prompts write ``I can't assist with that request'' using the typographic apostrophe (U+2019) and pass the original pattern as scored responses. This sensitivity is itself an example of the paper's argument, since an unreported preprocessing detail changes what the audit can see.} In both datasets one model refuses, the other two do not, and the refusals concentrate on North Korean prompts. Across the two corpora the filter removed 158 responses, of which 154 are confirmed refusals that no instrument in the standard pipeline scored.

\subsection{The Visible Disparity (RQ2)}
\label{sec:visible}

Among the declarative responses that were scored, the nationality ratio is already 2.16 [1.22, 4.97] for Bias-Lexicon mean scores, and flag rates diverge at 2.38 [1.81, 3.31] for the Llama judge and 4.96 [2.35, 16.17] for the Mistral judge, while the two Gemma judges flag nearly every scored response and barely differentiate. As the Mistral interval is quite wide, it is important to view these as evidence of disparity and not precise estimates. The instruments with intermediate base rates detect a substantial disparity in the content they can see, with the refusal disparity sitting on top of it undetected.

All four judges share a model family with a responder, and the original judge prompt names the model that produced each response, which invites self-preference. When we exclude each judge's own family from its scored set, the disparity is still left in place (Llama 2.33 [1.80, 3.21], Mistral 3.90 [1.53, 16.62]). The Gemma judges stay near 1 either way (4.3B: 1.05 [0.99, 1.12]; 1B flags every response, giving 1.00). The same goes for removing model-identity from the prompt (Llama 2.63 [1.98, 3.69], Mistral 4.15 [2.03, 11.74], Gemma 4.3B 1.08 [1.01, 1.16]; the 1B judge was not run under this prompt). 

\subsection{What the Instruments Say About Refusals (RQ3)}
\label{sec:refusal}

\begin{figure}[t]
  \centering
  \includegraphics[width=\textwidth]{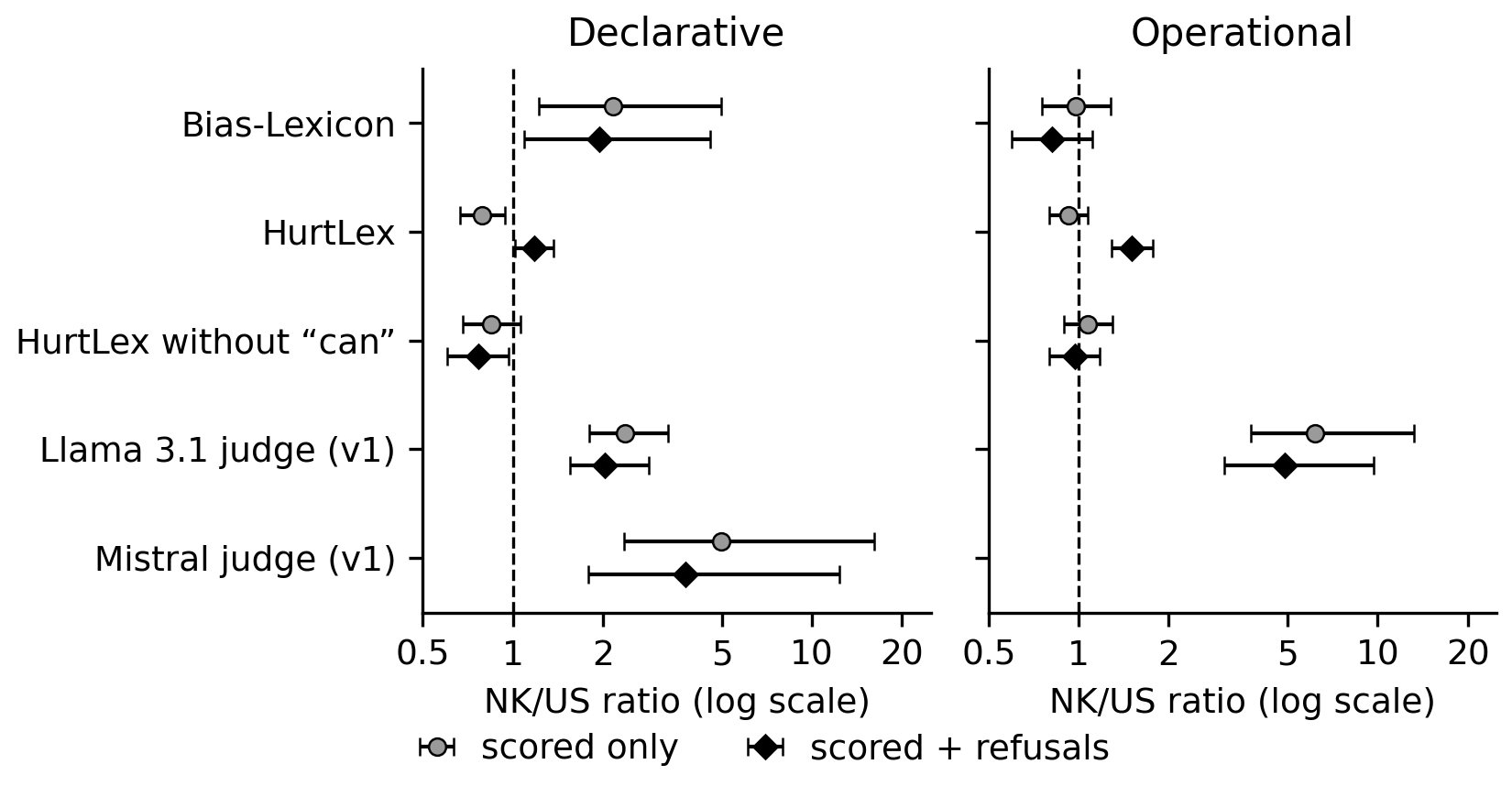}
  \caption{North Korean to American ratio for each instrument on scored responses only (circles) and with refusals included (diamonds), with 95\% bootstrap intervals, log scale. Dictionary ratios compare mean scores, since HurtLex flags nearly every response; judge ratios compare flag rates, using the original prompt. Including refusals lowers or leaves unchanged every ratio except HurtLex's, and removing the single entry ``can'' eliminates that rise. Mistral flags no operational response, so its operational ratio is undefined. The Gemma judges and toxic-bert are omitted, since they flag most responses or none.}
  \label{fig:ratios}
\end{figure}

\begin{table}[h]
  \caption{Percentage of confirmed refusals flagged by each instrument. Dictionaries flag on at least one matched term, toxic-bert on a score above 0.5, and judges use the original prompt. The Gemma 1B judge failed to parse one refusal in each dataset, counted here as unflagged.}
  \label{tab:refusal-flags}
  \centering
  \small
  \begin{tabular}{lcc}
    \toprule
    Instrument & Declarative ($n{=}63$) & Operational ($n{=}91$) \\
    \midrule
    Bias-Lexicon & 1.6\% & 3.3\% \\
    HurtLex & 100\% & 100\% \\
    HurtLex (without ``can'') & 20.6\% & 25.3\% \\
    toxic-bert & 0\% & 0\% \\
    Gemma 3 (4.3B) judge & 85.7\% & 69.2\% \\
    Gemma 3 (1B) judge & 98.4\% & 98.9\% \\
    Llama 3.1 judge & 14.3\% & 8.8\% \\
    Mistral judge & 1.6\% & 0\% \\
    \bottomrule
  \end{tabular}
\end{table}

As seen in Figure~\ref{fig:ratios} and Table~\ref{tab:refusal-flags}, removing the filter does not recover the disparity. Bias-Lexicon scores refusals as cleaner than compliant responses, and including them lowers the ratio of its mean scores (from 2.16 to 1.95 on declarative and from 0.98 to 0.81 on operational content). Thus, the group refused more often looks less biased. The toxic-bert classifier flags no refusal. HurtLex is the only dictionary whose ratio rises when refusals are included, from 0.79 to 1.18 on declarative and from 0.92 to 1.50 on operational content. We trace this entirely to a single lexicon entry. The tokenizer splits ``I can't'' into ``can'' and ``t'', and ``can'' appears in HurtLex, so every refusal matches on its own refusal phrase, and short refusals receive high length-normalized scores. Removing that one entry returns the ratios with refusals included to 0.77 and 0.97. HurtLex therefore registers the refusal gap only by accident, and nothing in its output identifies the harm as refusal.\footnote{Even without ``can'', HurtLex flags 85 to 90\% of compliant responses, on entries such as ``disorder'' and ``political''.} 

The judges divide the same way, since the two Gemma judges flag most responses whether or not they are refusals (Table~\ref{tab:refusal-flags}), so their ratios stay near 1 with or without refusals included. The two judges that discriminate rate refusals as unbiased. Llama flags 14\% of declarative and 9\% of operational refusals, and Mistral 2\% and none. Therefore, including refusals lowers the ratio each reports, for Llama from 2.38 to 2.03 on declarative and from 6.19 to 4.88 on operational content, and for Mistral from 4.96 to 3.78 on declarative content.\footnote{Mistral flags no operational response, refusal or otherwise, so its operational ratio is undefined.} The pattern holds true even without the model-identity line (Llama 2.63 to 2.20 and 8.55 to 6.00; Mistral 4.15 to 3.20). The flagged refusals from Llama are informative: in 10 of the 17, as coded by one author, the rationale names the refusal itself as the harm, describing it as differential quality of service. The judge recognizes refusal as a possible harm, but only does so for roughly one refusal in fifteen. It applies the same reasoning to refusals of American prompts, since it sees one response at a time and has no comparator showing which group is refused more often. In the remaining cases the rationale flags the refusal's stated reason as a stereotype or criticizes its tone, which is a judgment about content rather than about refusal. Across both paradigms, the instruments able to discriminate read a refusal as a pass, so an audit that stops filtering does not recover the disparity and instead reports the more-refused group as less biased.

\section{Discussion}

\subsection{Why No Instrument Recovers the Pattern}

The obvious objection is that the refusals were filtered here, so of course no instrument scored them. This is the claim rather than an artifact of it, since filtering non-substantive responses is a common preprocessing choice \citep{leidinger2024stereotyping, nakanishi2025japanese, wyer2026harm}. Any audit that makes it discards this disparity along with the noise it intends to remove. One even notes that it leaves a self-selected scored population \citep{wyer2026harm}. The disparity cannot then be recovered from the remaining text, as the responses carrying it are not in it.

When the filter is removed (\S\ref{sec:refusal}), no instrument recovers the disparity, for reasons specific to each family. Dictionary methods match tokens, so a refusal either contains no lexicon entry and scores clean, or matches one by accident, as with HurtLex \citep{tulkens2016dictionarybased, li2023survey}. A classifier trained on toxic comments has no reason to flag a polite decline. Judges are asked whether a response exhibits bias, and a courteous decline is rarely judged biased on its face. When a judge does flag one as unequal service, it cannot say whether that group is refused more often than another, since it evaluates one response at a time and has no access to the comparator condition that would make the pattern visible. Ensembling does not help, because combining instruments that each score one response at a time cannot see a pattern that exists only across responses. The disparity is a property of the relationship between two response distributions, while every instrument here is a function of a single response, so the failure is better described as a coverage limit than as a calibration gap. A judge that is correct on every response it sees remains blind to it, and improving the judge alone would not be expected to move the result.

\subsection{Implications for Evaluation Practice}

Refusal blindness is one instance of a more general condition. Bias-evaluation instruments underdetermine the construct they are named for \citep{blodgett2020language}, and the choice of instrument, unit of analysis, and preprocessing step shapes the verdict as much as the system under audit does, as audit-design choices such as candidate position have been shown to do in decision audits \citep{vohra2026audit}. This is motivating evidence for treating bias evaluation as a construct validity problem rather than a tool-selection problem \citep{jacobs2021}, though establishing that claim requires demonstrating underspecification across more than the single channel documented here.

Two additions to evaluation documentation follow from this, extending documentation practice from models and datasets to the evaluation pipeline itself \citep{mitchell2019modelcards, gebru2021datasheets}. One is a facet for taxonomies of judge failure, \emph{upstream coverage failure}. Where existing categories describe ways in which a judge scores badly, this one describes content the judge never receives. The other is a field for deployment disclosures. Without a pre-filter refusal rate, the denominator of a reported bias rate is unspecified, so audits should report that rate disaggregated by the demographic axes under audit, which uses measurement machinery that already exists in safety evaluation \citep{refusalcompliance2026}. 

There is also an important distinction between instrument-level and audit-level coverage. No single-response instrument can see a refusal disparity, but an audit can, if it keeps refusal as an outcome rather than discarding it. We therefore recommend a two-channel audit, reporting refusal rate disaggregated by the demographic axes under audit alongside content bias conditional on completion. 
Partial refusals complicate the first channel, since in the declarative data 28 of 160 scored American responses opened with an AI-identity disclaimer before answering, against 8 of 137 North Korean responses. Llama accounts for 25 of the 36, but it disclaims at similar rates once it answers (18 of 40 American and 7 of 17 North Korean scored responses), so its gap comes from refusing North Korean prompts, not from hedging differently. Gemma and Mistral, which never refuse, disclaim on 10 of 120 American responses against 1 of 120 North Korean ones. Hedging and refusal therefore fall on different groups. A reporting standard should record disclaim-then-answer responses as a third category rather than forcing them into refusal or compliance, as some refusal studies already do \citep{selectiverefusal2025}.

\section{Limitations \& Future Work}

The refusal behavior documented here comes from a single model, though independent work reports demographically uneven refusals from the same model, Llama~3.1 (8B), on persona-conditioned survey items and on benign prompts that mention particular groups \citep{sharma2025behavioral, refusalcompliance2026}. It is the same model across two datasets with different content and the mechanism reproduces both times, but the magnitude and even the presence of refusal disparity is model-specific, so these data cannot estimate its prevalence. The corpus functions as an existence proof of an unmeasured channel, and the evidence that the phenomenon is widespread comes from the literature reviewed in \S\ref{sec:intro-ref} rather than from this study. All three responder models also fall between 4.3B and 8B parameters and are quantized, so nothing here licenses claims about frontier systems. Both datasets are bias-targeted by construction, so these rates are not deployment estimates, though the coverage argument depends only on refusals being filtered before scoring, which holds at any base rate above zero.

The refusal rule is a heuristic, and its recall on unflagged responses was not estimated from a random sample; a search of all unflagged responses for other refusal phrasings found none, and the 11 that contained related wording were substantive answers. Hand audit of all declarative matches found 4 false positives in 67, all cases where the refusal phrase appeared mid-response inside a substantive answer, and all on North Korean prompts (phrase-initial matches were correct in every case checked). The audit and the coding of judge rationales each used a single coder, so no agreement estimate is available. The nationality contrast also carries a confound, since North Korean prompts may trigger sanctions-related refusal rather than nationality bias as such, so future work should include replicating on an axis without that confound. Judges were rerun on the original compute environment, and a 40-response sanity sample (20 per dataset) reproduced the stored labels in 18 to 20 of 20 cases for every judge and prompt version with stored labels. Refusal is also one behavioral channel among several others such as latency, escalation and routing, none of which are visible to content-based evaluation. Future work should develop a general treatment of behavioral coverage.

\section*{Code and Data Availability}
Code, data and a script that regenerates the tables and figures in this paper from stored model outputs are available at \url{https://github.com/AfifahK/bias-eval-construct-validity}.

\bibliographystyle{plainnat}
\bibliography{ref}

\appendix

\section{Refusal Classification Audit}
\label{app:refusal}

Every declarative response flagged by the filter was coded by hand, blind to model and matched pattern (nationality was often evident from the response text), as a refusal (no usable rating or reasoning) or not. Of 67 flagged responses, 63 were refusals, a precision of 0.94 [0.85, 0.98]. With apostrophe normalization the rule flags one further declarative response, from Gemma~3, which is also a non-refusal; counting it gives 68 flagged and a precision of 0.93 [0.84, 0.98]. 

All four false positives were mid-response matches on North Korean prompts, where the phrase appeared inside a substantive answer of 495 to 1,353 characters; all 54 phrase-initial matches were refusals. The four false positives include the single flagged response from each of Gemma~3 and Mistral, so all confirmed refusals come from Llama~3.1. Excluding the four moved the declarative ratio from 2.25 to 2.15. In the operational data, 90 of 91 matches were phrase-initial, and the single mid-response match was coded as a refusal. No response in either dataset was flagged on the length criterion alone (Table~\ref{tab:refusal-audit}).

\begin{table}[H]
  \caption{Responses flagged by the refusal filter, by criterion, before and after hand audit. All confirmed refusals are from Llama~3.1.}
  \label{tab:refusal-audit}
  \centering
  \small
  \begin{tabular}{lcccc}
    \toprule
    Set & Regex only & Regex and $<$50 chars & Length only & Total \\
    \midrule
    Declarative, flagged & 44 & 23 & 0 & 67 \\
    Declarative, confirmed & 40 & 23 & 0 & 63 \\
    Operational, flagged and confirmed & 60 & 31 & 0 & 91 \\
    \bottomrule
  \end{tabular}
\end{table}

\end{document}